# GEMINI: A NATURAL LANGUAGE SYSTEM FOR SPOKEN-LANGUAGE UNDERSTANDING[*]

John Dowding, Jean Mark Gawron, Doug Appelt,
John Bear, Lynn Cherny, Robert Moore, and Douglas Moran

SRI International
333 Ravenswood Avenue
Menlo Park, CA 94025
Internet: dowding@ai.sri.com [†]

## 1. ABSTRACT

Gemini is a natural language understanding system developed for spoken language applications. The paper describes the architecture of Gemini, paying particular attention to resolving the tension between robustness and overgeneration. Gemini features a broad-coverage unification-based grammar of English, fully interleaved syntactic and semantic processing in an all-paths, bottom-up parser, and an utterance-level parser to find interpretations of sentences that might not be analyzable as complete sentences. Gemini also includes novel components for recognizing and correcting grammatical disfluencies, and for doing parse preferences. This paper presents a component-by-component view of Gemini, providing detailed relevant measurements of size, efficiency, and performance.

## 2. INTRODUCTION

Gemini is a natural language (NL) understanding system developed for spoken language applications. This paper describes the details of the system, and includes relevant measurements of size, efficiency, and performance of each of its components.

In designing any NL understanding system, there is a tension between robustness and correctness. Forgiving an error risks throwing away crucial information; furthermore, devices added to a system to enhance robustness can sometimes enrich the ways of finding an analysis, multiplying the number of analyses for a given input, and making it more difficult to find the correct analysis. In processing spoken language this tension is heightened because the task of speech recognition introduces a new source of error. The robust system will attempt to find a sensible interpretation, even in the presence of performance errors by the speaker, or recognition errors by the speech recognizer. On the other hand, a system should be able to detect that a recognized string is not a sentence of English, to help filter recognition errors by the speech recognizer. Furthermore, if parsing and recognition are interleaved, then the parser should enforce constraints on partial utterances.

The approach taken in Gemini is to constrain language recognition with fairly conventional grammar, but to augment that grammar with two orthogonal rule-based recognition modules, one for glueing together the fragments found during the conventional grammar parsing phase, and another for recognizing and eliminating disfluencies known as "repairs." At the same time, the multiple analyses arising before and after all this added robustness are managed in two ways: first, by highly constraining the additional rule-based modules by partitioning the rules into preference classes, and second, through the addition of a postprocessing parse preference component.

Processing starts in Gemini when syntactic, semantic, and lexical rules are applied by a bottom-up all-paths *constituent* parser to populate a chart with edges containing syntactic, semantic, and logical form information. Then, a second *utterance* parser is used to apply a second set of syntactic and semantic rules that are required to span the entire utterance. If no semantically acceptable utterance-spanning edges are found during this phase, a component to recognize and correct certain grammatical disfluencies is applied. When an acceptable interpretation is found, a set of parse preferences is used to choose a single best interpretation from the chart to be used for subsequent processing. Quantifier scoping rules are applied to this best interpretation to produce the final logical form, which is then used as input to a query-answering system. The following sections describe each of these components in detail, with the exception of the query-answering subsystem, which is not described in this paper.

[*]This research was supported by the Advanced Research Projects Agency under Contract ONR N00014-90-C-0085 with the Office of Naval Research. The views and conclusions contained in this document are those of the authors and should not be interpreted as necessarily representing the official policies, either expressed or implied, of the Advanced Research Projects Agency of the U.S. Government.

[†]In 'Proceedings of the 31[st] Annual Meeting of the Association for Computational Lingistics,' June 1993.

In our component-by-component view of Gemini, we provide detailed statistics on each component's size, speed, coverage, and accuracy. These numbers detail our performance on the sub-domain of air-travel planning that is currently being used by the ARPA spoken language understanding community (MADCOW, 1992). Gemini was trained on a 5875-utterance dataset from this domain, with another 688 utterances used as a blind test (not explicitly trained on, but run multiple times) to monitor our performance on a dataset on which we did not train. We also report here our results on another 756-utterance fair test set that we ran only once. Table 1 contains a summary of the coverage of the various components on both the training and fair test sets. More detailed explanations of these numbers are given in the relevant sections.

|  | Training | Test |
|---|---|---|
| Lexicon | 99.1% | 95.9% |
| Syntax | 94.2% | 90.9% |
| Semantics | 87.4% | 83.7% |
| Syntax (repair correction) | 96.0% | 93.1% |
| Semantics (repair correction) | 89.1% | 86.0% |

Table 1: Domain Coverage by Component

## 3. SYSTEM DESCRIPTION

Gemini maintains a firm separation between the language- and domain-specific portions of the system, and the underlying infrastructure and execution strategies. The Gemini kernel consists of a set of compilers to interpret the high-level languages in which the lexicon and syntactic and semantic grammar rules are written, as well as the parser, semantic interpretation, quantifier scoping, repair correction mechanisms, and all other aspects of Gemini that are not specific to a language or domain. Although this paper describes the lexicon, grammar, and semantics of English, Gemini has also been used in a Japanese spoken language understanding system (Kameyama, 1992).

### 3.1. Grammar Formalism

Gemini includes a midsized constituent grammar of English (described in section 2.3), a small utterance grammar for assembling constituents into utterances (described in section 2.7), and a lexicon. All three are written in a variant of the unification formalism used in the Core Language Engine (Alshawi, 1992) .

The basic building block of the grammar formalism is a category with feature constraints. Here is an example:
```
np:[wh=ynq,case=(nom∨acc),
    pers_num=(3rd∧sg)]
```
This category can be instantiated by any noun phrase with the value ynq for its wh feature (which means it must be a wh-bearing noun phrase like *which book*, *who*, or *whose mother*), either acc (accusative) or nom (nominative) for its case feature, and the conjunctive value 3rd∧sg (third and singular) for its person-number feature. This formalism is related directly to the Core Language Engine, but more conceptually it is closely related to that of other unification-based grammar formalisms with a context-free skeleton, such as PATR-II (Shieber et al., 1983), Categorial Unification Grammar (Uszkoreit, 1986), Generalized Phrase-Structure Grammar (Gazdar et al., 1982), and Lexical Functional Grammar (Bresnan, 1982).

Gemini differs from other unification formalisms in the following ways. Since many of the most interesting issues regarding the formalism concern typing, we defer discussing motivation until section 2.5.

- Gemini uses typed unification. Each category has a set of features declared for it. Each feature has a declared value space of possible values (value spaces may be shared by different features). Feature structures in Gemini can be recursive, but only by having categories in their value space; so typing is also recursive. Typed feature structures are also used in HPSG (Pollard and Sag, in press). One important difference with the use in Gemini is that Gemini has no type inheritance.

- Some approaches do not assume a *syntactic skeleton* of category-introducing rules (for example, Functional Unification Grammar (Kay, 1979)). Some make such rules implicit (for example, the various categorial unification approaches, such as Unification Categorial Grammar (Zeevat, Klein, and Calder, 1987)).

- Even when a syntactic skeleton is assumed, some approaches do not distinguish the category of a constituent (for example, np, vp) from its other features (for example, pers_num, gapsin, gapsout). Thus, for example, in one version of GPSG, categories were simply feature bundles (attribute value structures) and there was a feature MAJ taking values like N,V,A, and P which determined the major category of constituent.

- Gemini does not allow rules schematizing over syntactic categories.

### 3.2. Lexicon

The Gemini lexicon uses the same category notation as the Gemini syntactic rules. Lexical categories are types as well, with sets of features defined for them. The lexical component of Gemini includes the lexicon of base forms, lexical templates, morphological rules, and the lexical type and feature default specifications.

The Gemini lexicon used for the air-travel planning domain contains 1,315 base entries. These expand by morphological rules to 2,019. In the 5875-utterance training set, 52 sentences contained unknown words (0.9%), compared to 31 sentences in the 756-utterance fair test set (4.1%).

### 3.3. Constituent Grammar

A simplified example of a syntactic rule is

```
syn(whq_ynq_slash_np,
    [ s:[sentence_type=whq, form=tnsd,
         gapsin=G, gapsout=G],
      np:[wh=ynq, pers_num=N],
      s:[sentence_type=ynq, form=tnsd,
         gapsin=np:[pers_num=N],
         gapsout=null]]).
```

This syntax rule (named whq_ynq_slash_np) says that a sentence (category s) can be built by finding a noun phrase (category np) followed by a sentence. It requires that the daughter np have the value ynq for its wh feature and that it have the value N (a variable) for its person-number feature. It requires that the daughter sentence have a category value for its gapsin feature, namely an np with a person number value N, which is the same as the person number value on the wh-bearing noun phrase. The interpretation of the entire rule is that a gapless sentence with sentence_type whq can be built by finding a wh-phrase followed by a sentence with a noun phrase gap in it that has the same person number as the wh-phrase.

Semantic rules are written in much the same rule format, except that in a semantic rule, each of the constituents mentioned in the phrase structure skeleton is associated with a logical form. Thus, the semantics for the rule above is

```
sem(whq_ynq_slash_np,
    [([whq,S], s:[]),
     (Np, np:[]),
     (S, s:[gapsin=np:[gapsem=Np]])]).
```

Here the semantics of the mother s is just the semantics of the daughter s with the illocutionary force marker whq wrapped around it. In addition, the semantics of the s gap's np's gapsem has been unified with the semantics of the wh-phrase. Through a succession of unifications this will end up assigning the wh-phrase's semantics to the gap position in the argument structure of the s. Although each semantic rule must be keyed to a preexisting syntactic rule, there is no assumption of rule-to-rule uniqueness. Any number of semantic rules may be written for a single syntactic rule. We discuss some further details of the semantics in section 2.6

The constituent grammar used in Gemini contains 243 syntactic rules, and 315 semantic rules. Syntactic coverage on the 5875-utterance training set was 94.2%, and on the 756-utterance test set it was 90.9%.

### 3.4. Parser

Since Gemini was designed with spoken language interpretation in mind, key aspects of the Gemini parser are motivated by the increased needs for robustness and efficiency that characterize spoken language. Gemini uses essentially a pure bottom-up chart parser, with some limited left-context constraints applied to control creation of categories containing syntactic gaps.

Some key properties of the parser are

- The parser is all-paths bottom-up, so that all possible edges admissible by the grammar are found.

- The parser uses subsumption checking to reduce the size of the chart. Essentially, an edge is not added to the chart if it is less general than a preexisting edge, and preexisting edges are removed from the chart if the new edge is more general.

- The parser is *on-line* (Graham, Harrison, and Russo, 1980), essentially meaning that all edges that end at position $i$ are constructed before any that end at position $i+1$. This feature is particularly desirable if the final architecture of the speech understanding system couples Gemini tightly with the speech recognizer, since it guarantees for any partial recognition input that all possible constituents will be built.

An important feature of the parser is the mechanism used to constrain the construction of categories containing syntactic gaps. In earlier work (Moore and Dowding, 1991), we showed that approximately 80% of the edges built in an all-paths bottom-up parser contained gaps, and that it is possible to use prediction in a bottom-up parser only to constrain the gap categories, without requiring prediction for nongapped categories. This limited form of left-context constraint greatly reduces the total number of edges built for a very low overhead. In the 5875-utterance training set, the chart for the average sentence contained 313 edges, but only 23 predictions.

### 3.5. Typing

The main advantage of typed unification is for grammar development. The type information on features allows the lexicon, grammar, and semantics compilers to provide detailed error analysis regarding the flow of values through the grammar, and to warn if features are assigned improper values, or variables of incompatible types are unified. Since the type-analysis is performed statically at compile time, there is no run-time overhead associated with adding types to the grammar.

The major grammatical category plays a special role in the typing scheme of Gemini. For each category, Gemini makes a set of declarations stipulating its allowable features and the relevant value spaces. Thus, the distinction between the syntactic category of a constituent and its other features can be cashed out as follows: the syntactic category can be thought of as the feature structure type. The only other types needed by Gemini are the value spaces used by features. Thus for example, the type v (verb) admits a feature vform, whose value space vform-types can be instantiated with values like present participle, finite, and past participle. Since all recursive features are category-valued, these two kinds of types suffice.

### 3.6. Interleaving Syntactic and Semantic Information

**Sortal Constraints** Selectional restrictions are imposed in Gemini through the sorts mechanism. Selectional restrictions include both highly domain-specific information about predicate-argument and very general predicate restrictions. For example, in our application the object of the transitive verb *depart* (as in *flights departing Boston*) is restricted to be an airport or a city, obviously a domain-specific requirement. But the same machinery also restricts a determiner like *all* to take two propositions, and an adjective like *further* to take distances as its measure-specifier (as in *thirty miles further*). In fact, sortal constraints are assigned to every atomic predicate and operator appearing in the logical forms constructed by the semantic rules.

Sorts are located in a conceptual hierarchy and are implemented as Prolog terms such that more general sorts subsume more specific sorts (Mellish, 1988). This allows the subsumption checking and packing in the parser to share structure whenever possible. Semantic coverage with sortal constraints applied was 87.4% on the training set, and on the test set it was 83.7%.

**Interleaving Semantics with Parsing** In Gemini, syntactic and semantic processing is fully interleaved. Building an edge requires that syntactic constraints be applied, which results in a tree structure, to which semantic rules can be applied, which results in a logical form to which sortal constraints can be applied. Only if the syntactic edge leads to a well-sorted semantically-acceptable logical form fragment is it added to the chart.

|  | Edges | Time |
|---|---|---|
| Syntax only | 197 | 3.4 sec. |
| Syntax + semantics | 234 | 4.47 sec. |
| Syntax + semantics + sorts | 313 | 13.5 sec. |

Table 2: Average Number of Edges Built by Interleaved Processing

Interleaving the syntax and semantics in this way depends on a crucial property of the semantics: a semantic interpretation is available for each syntactic node. This is guaranteed by the semantic rule formalism and by the fact that every lexical item has a semantics associated with it.

Table 2 contains average edge counts and parse timing statistics[1] for the 5875-utterance training set.

### 3.7. Utterance Parsing

The constituent parser uses the constituent grammar to build all possible categories bottom-up, independent of location within the string. Thus, the constituent parser does not force any constituent to occur either at the beginning of the utterance, or at the end. Those constraints are stated in what we call the utterance grammar. They are applied after constituent parsing is complete by the *utterance* parser. The utterance grammar specifies ways of combining the categories found by the constituent parser into an analysis of the complete utterance. It is at this point that the system recognizes whether the sentence was a simple complete sentence, an isolated sentence fragment, a run-on sentence, or a sequence of related fragments.

Many systems (Carbonell and Hayes, 1983), (Hobbs et al., 1992), (Seneff, 1992), (Stallard and Bobrow, 1992) have added robustness with a similar postprocessing phase. The approach taken in Gemini differs in that the utterance grammar uses the same syntactic and semantic rule formalism used by the constituent grammar. Thus, the same kinds of logical forms built during constituent parsing are the output of utterance parsing, with the same sortal constraints enforced. For example, an utterance consisting of a sequence of modifier fragments (like *on Tuesday at three o'clock on United*) is interpreted as a conjoined property of a flight, because the only sort of thing in the ATIS domain that can be on Tuesday at three o'clock on United is a flight.

---
[1] Gemini is implemented primarily in Quintus Prolog version 3.1.1. All timing numbers given in this paper were run on a lightly loaded Sun SPARCstation 2 with at least 48 MB of memory. Under normal conditions, Gemini runs in under 12 MB of memory.

The utterance parser partitions the utterance grammar into equivalence classes and considers each class according to an ordering. Utterance parsing terminates when all constituents satisfying the rules of the current equivalence class are built, unless there are none, in which case the next class is considered. The highest ranked class consists of rules to identify simple complete sentences, the next highest class consists of rules to identify simple isolated sentence fragments, and so on. Thus, the utterance parser allows us to enforce a very coarse form of parse preferences (for example, prefering complete sentences to sentence fragments). These coarse preferences could also be enforced by the parse preference component described in section 2.9, but for the sake of efficiency we choose to enforce them here.

The utterance grammar is significantly smaller than the constituent grammar – only 37 syntactic rules and 43 semantic rules.

### 3.8. Repairs

Grammatical disfluencies occur frequently in spontaneous spoken language. We have implemented a component to detect and correct a large subclass of these disfluencies (called repairs, or self-corrections) where the speaker intends that the meaning of the utterance be gotten by deleting one or more words. Often, the speaker gives clues of their intention by repeating words or adding cue words that signal the repair:

(1) a. How many American airline flights leave Denver on June June tenth.
  b. Can you give me information on all the flights from San Francisco no from Pittsburgh to San Francisco on Monday.

The mechanism used in Gemini to detect and correct repairs is currently applied as a fallback if no semantically acceptable interpretation is found for the complete utterance. The mechanism finds sequences of identical or related words, possibly separated by a cue word (for example, oh or no) that might indicate the presence of a repair, and deletes the first occurrence of the matching portion. Since there may be several such sequences of possible repairs in the utterance, the mechanism produces a ranked set of candidate corrected utterances. These candidates are ranked in order of the fewest deleted words. The first candidate that can be given an interpretation is accepted as the intended meaning of the utterance. This approach is presented in detail in (Bear, Dowding, and Shriberg, 1992).

The repair correction mechanism helps increase the syntactic and semantic coverage of Gemini (as reported in Table 1). In the 5875-utterance training set, 178 sentences contained nontrivial repairs[2], of which Gemini found 89 (50%). Of the sentences Gemini corrected, 81 were analyzed correctly (91%), and 8 contained repairs but were corrected wrongly. Similarly, the 756-utterance test set contained 26 repairs, of which Gemini found 11 (42%). Of those 11, 8 were analyzed correctly (77%), and 3 were analyzed incorrectly.

Since Gemini's approach is to extend language analysis to recognize specific patterns characteristic of spoken language, it is important for components like repair correction (which provide the powerful capability of deleting words) not to be applied in circumstances where no repair is present. In the 5875-utterance training set, Gemini misidentified only 15 sentences (0.25%) as containing repairs when they did not. In the 756-utterance test set, only 2 sentences were misidentified as containing repairs (0.26%).

While the repair correction component currently used in Gemini does not make use of acoustic/prosodic information, it is clear that acoustics can contribute meaningful cues to repair. In future work, we hope to improve the performance of our repair correction component by incorporating acoustic/prosodic techniques for repair detection (Bear, Dowding, and Shriberg, 1992) (Nakatani and Hirschberg, 1993) (O'Shaughnessy, 1992).

A central question about the repairs module concerns its role in a tightly integrated system in which the NL component filters speech recognition hypotheses. The open question: should the repairs module be part of the recognizer filter or should it continue to be a post-processing component? The argument for including it in the filter is that without a repairs module, the NL system rejects many sentences with repairs, and will thus disprefer essentially correct recognizer hypotheses. The argument against including it is efficiency and the concern that with recognizer errors present, the repair module's precision may suffer: it may attempt to repair sentences with no repair in them. Our current best guess is that recognizer errors are essentially orthogonal to repairs and that a filter including the repairs module will not suffer from precision problems. But we have not yet performed the experiments to decide this.

---

[2]For these results, we ignored repairs consisting of only an isolate fragment word, or sentence-initial filler words like "yes" and "okay".

### 3.9. Parse Preference Mechanism

In Gemini, parse preferences are enforced when *extracting* syntactically and semantically well-formed parse trees from the chart. In this respect, our approach differs from many other approaches to the problem of parse preferences, which make their preference decisions as parsing progresses, pruning subsequent parsing paths (Frazier and Fodor, 1978), (Hobbs and Bear, 1990), (Marcus 1980). Applying parse preferences requires comparing two subtrees spanning the same portion of the utterance.

The parse preference mechanism begins with a simple strategy to disprefer parse trees containing specific "marked" syntax rules. As an example of a dispreferred rule, consider: *Book those three flights to Boston.* This sentence has a parse on which *those three* is a noun phrase with a missing head (consider a continuation of the discourse *Three of our clients have sufficient credit*). After penalizing such dispreferred parses, the preference mechanism applies attachment heuristics based on the work by Pereira (1985) and Shieber (1983)

Pereira's paper shows how the heuristics of Minimal Attachment and Right Association (Kimball, 1973) can both be implemented using a bottom-up shift-reduce parser.

(2)(a) John sang a song for Mary.
   (b) John canceled the room Mary reserved yesterday.

Minimal Attachment selects for the tree with the fewest nodes, so in (2a), the parse that makes *for Mary* a complement of *sings* is preferred. Right Association selects for the tree that incorporates a constituent A into the rightmost possible constituent (where rightmost here means *beginning the furthest to the right*). Thus, in (2b) the parse in which *yesterday* modifies *reserved* is preferred.

The problem with these heuristics is that when they are formulated loosely, as in the previous paragraph, they appear to conflict. In particular, in (2a), Right Association seems to call for the parse that makes *for Mary* a modifier of *song*.

Pereira's goal is to show how a shift-reduce parser can enforce both heuristics without conflict and enforce the desired preferences for examples like (2a) and (2b). He argues that Minimal Attachment and Right Association can be enforced in the desired way by adopting the following heuristics for resolving conflicts:

1. Right Association: In a shift-reduce conflict, prefer shifts to reduces.
2. Minimal Attachment: In a reduce-reduce conflict, prefer longer reduces to shorter reduces.

Since these two principles never apply to the same choice, they never conflict.

For purposes of invoking Pereira's heuristics, the derivation of a parse can be represented as the sequence of S's (Shift) and R's (Reduce) needed to construct the parse's unlabeled bracketing. Consider, for example, the choice between two unlabeled bracketings of (2a):

(a)  [John [sang [a song ] [for Mary ] ] ]
      S    S    SS   R S   S     RRR
(b)  [John [sang [ [a song ] [for Mary ] ] ] ]
      S    S      SS   R S   S      RRRR

There is a shift for each word and a reduce for each right bracket. Comparison of the two parses consists simply of pairing the moves in the shift-reduce derivation from left to right. Any parse making a shift move that corresponds to a reduce move loses by Right Association. Any parse making a reduce move that corresponds to a longer reduce loses by Minimal Attachment. In derivation (b) above, the third reduce move builds the constituent *a song for Mary* from two constituents, while the corresponding reduce in (a) builds *sang a song for Mary* from three constituents. Parse (b) thus loses by Minimal Attachment.

Questions about the exact nature of parse preferences (and thus about the empirical adequacy of Pereira's proposal) still remain open, but the mechanism sketched does provide plausible results for a number of examples.

### 3.10. Scoping

The final logical form produced by Gemini is the result of applying a set of quantifier scoping rules to the best interpretation chosen by the parse preference mechanism. The semantic rules build *quasi-logical forms*, which contain complete semantic predicate-argument structure, but do not specify quantifier scoping. The scoping algorithm that we use combines syntactic and semantic information with a set of quantifier scoping preference rules to rank the possible scoped logical forms consistent with the quasi-logical form selected by parse preferences. This algorithm is described in detail in (Moran, 1988).

### 4. CONCLUSION

In our approach to resolving the tension between overgeneration and robustness in a spoken language understanding system, some aspects of Gemini are specifically oriented towards limiting overgeneration, such as the on-line property for the parser, and fully interleaved syntactic and semantic processing. Other components, such as the fragment and run-on processing provided by the utterance grammar, and the correction of recognizable grammatical repairs, increase the robustness of Gemini. We believe a robust system can still recognize and disprefer utterances containing recognition errors.

Research in the construction of the Gemini system is ongoing to improve Gemini's speed and coverage, as well as to examine deeper integration strategies with speech recognition, and integration of prosodic information into spoken language disambiguation.

## REFERENCES

Alshawi, H. (ed) (1992). *The Core Language Engine*, MIT Press, Cambridge.

Bear, J., Dowding, J., and Shriberg, E. (1992). "Integrating Multiple Knowledge Sources for the Detection and Correction of Repairs in Human-Computer Dialog", in *Proceedings of the 30th Annual Meeting of the Association for Computational Linguists*, Newark, DE, pp. 56-63.

Bresnan, J. (ed) (1982). *The Mental Representation of Grammatical Relations*, MIT Press, Cambridge.

Carbonell, J., and Hayes, P. (1983). "Recovery Strategies for Parsing Extragrammatical Language", *American Journal of Computational Linguistics*, Vol. 9, Numbers 3-4, pp. 123-146.

Frazier, L., and Fodor, J.D. (1978). "The Sausage Machine: A New Two-Stage Parsing Model", *Cognition*, Vol. 6, pp. 291-325.

Gazdar, G., Klein, E., Pullum, G., and Sag, I. (1982). *Generalized Phrase Structure Grammar*, Harvard University Press, Cambridge.

Graham, S., Harrison, M., and Ruzzo, W. (1980). "An Improved Context-Free Recognizer", *ACM Transactions on Programming Languages and Systems*, Vol. 2, No. 3, pp. 415-462.

Hobbs, J., and Bear, J. (1990). "Two Principles of Parse Preference", in *Proceedings of the 13th International Conference on Computational Linguistics*, Helsinki, Vol. 3, pp. 162-167.

Hobbs, J., Appelt, D., Bear, J., Tyson, M., and Magerman, D. (1992). "Robust Processing of Real-World Natural-Language Texts", in *Text Based Intelligent Systems*, ed. P. Jacobs, Lawrence Erlbaum Associates, Hillsdale, NJ, pp. 13-33.

Kameyama, M. (1992). "The Syntax and Semantics of the Japanese Language Engine", forthcoming. In *Mazuka, R., and N. Nagai, Eds. Japanese Syntactic Processing*, Hillsdale, NJ: Lawrence Erlbaum Associates.

Kay, M. (1979). "Functional Grammar", in *Proceedings of the 5th Annual Meeting of the Berkeley Linguistics Society.* pp. 142-158.

Kimball, J. (1973). "Seven Principles of Surface Structure Parsing in Natural Language", *Cognition*, Vol. 2, No. 1, pp. 15-47.

MADCOW (1992). "Multi-site Data Collection for a Spoken Language Corpus", in *Proceedings of the DARPA Speech and Natural Language Workshop*, February 23-26, 1992.

Marcus, M. (1980). *A Theory of Syntactic Recognition for Natural Language*, MIT Press, Cambridge.

Moran, D. (1988). "Quantifier Scoping in the SRI Core Language Engine", in *Proceedings of the 26th Annual Meeting of the Association for Computational Linguistics*, State University of New York at Buffalo, Buffalo, NY, pp. 33-40.

Mellish, C. (1988). "Implementing Systemic Classification by Unification". *Computational Linguistics* Vol. 14, pp. 40-51.

Moore, R., and Dowding, J. (1991). "Efficient Bottom-up Parsing", in *Proceedings of the DARPA Speech and Natural Language Workshop*, February 19-22, 1991, pp. 200-203.

Nakatani, C., and Hirschberg, J. (1993). "A Speech-First Model for Repair Detection and Correction", in *Proceedings of the ARPA Workshop on Human Language Technology*, March 21-24, 1993, Plainsboro, NJ.

O'Shaughnessy, D. (1992). "Analysis of False Starts in Spontaneous Speech", in *Proceedings of the 1992 International Conference on Spoken Language Processing*, October 12-16, 1992, Banff, Alberta, Canada, pp. 931-934.

Pereira, F. (1985). "A New Characterization of Attachment Preferences", in *Natural Language Parsing*, Ed. by Dowty, D., Karttunen, L., and Zwicky, A., Cambridge University Press, Cambridge, pp. 307-319.

Pollard, C., and Sag, I. (in press). *Information-Based Syntax and Semantics, Vol. 2*, CSLI Lecture Notes.

Seneff, S. (1992). "A Relaxation Method for Understanding Spontaneous Speech Utterances", in *Proceedings of the Speech and Natural Language Workshop*, Harriman, NY, pp. 299-304.

Shieber, S. (1983). "Sentence Disambiguation by a Shift-Reduce Parsing Technique", in *Proceedings of the 21 Annual Meeting of the Association for Computational Linguistics*, Boston, Massachusetts, pp. 113-118.

Shieber, S., Uszkoreit, H., Pereira, F., Robinson, J., and Tyson, M. (1983). "The Formalism

and Implementation of PATR-II", in Grosz, B. and Stickel, M. (eds) *Research on Interactive Acquisition and Use of Knowledge*, SRI International, pp. 39-79.

Stallard, D., and Bobrow, R. (1992). "Fragment Processing in the DELPHI System", in *Proceedings of the Speech and Natural Language Workshop*, Harriman, NY, pp. 305-310.

Uszkoreit, H. (1986). "Categorial Unification Grammars", in *Proceedings of the 11th International Conference on Computational Linguistics and the 24th Annual Meeting of the Association for Computational Linguistics*, Institut fur Kummunikkationsforschung und Phonetik, Bonn University.

Zeevat, H., Klein, E., and Calder, J. (1987). "An Introduction to Unification Categorial Grammar", in Haddock, N., Klein, E., Merrill, G. (eds.) *Edinburgh Working Papers in Cognitive Science, Volume 1: Categorial Grammar, Unification Grammar, and Parsing*.